# Color Texture Classification Based on Proposed Impulse-Noise Resistant Color Local Binary Patterns and Significant Points Selection Algorithm

*Shervan Fekri-Ershad[*] and Farshad Tajeripour*
Department of Computer Science and Engineering, Shiraz University, Shiraz, Iran
* fekriershad@pco.iaun.ac.ir

**Abstract**
**Purpose -** The main aim of this paper is to propose a color-texture classification approach which uses color sensor information and texture features jointly. High accuracy, low noise sensitivity and low computational complexity are specified aims for our proposed approach.
**Design/methodology/approach** – One of the efficient texture analysis operations is local binary patterns (LBP). The proposed approach includes two steps. First, a noise resistant version of color LBP is proposed to decrease it's sensitivity to noise. This step is evaluated based on combination of color sensor information using AND operation. In second step, a significant points selection (SPS) algorithm is proposed to select significant LBPs. This phase decreases final computational complexity along with increasing accuracy rate.
**Findings** - The Proposed approach is evaluated using Vistex, Outex, and KTH-TIPS-2a data-sets. Our approach has been compared with some state-of-the-art methods. It is experimentally demonstrated that the proposed approach achieves highest accuracy. In two other experiments, result show low noise sensitivity and low computational complexity of the proposed approach in comparison with previous versions of LBP. Rotation invariant, multi-resolution, general usability are other advantages of our proposed approach.
**Originality/value** – In the present paper, a new version of LBP is proposed originally, which is called Hybrid color local binary patterns (HCLBP). HCLBP can be used in many image processing applications to extract color/texture features jointly. Also, a significant point selection algorithm is proposed for the first time to select key points of images.

**Paper Type** Research paper

## 1. INTRODUCTION

A surface texture is created by the regular repetition of an element or pattern on a surface. One of the basic parameters of human vision by which we discriminate between surfaces and objects is texture analysis (Petrou and Sevilla, 2009). Visual sensor technologies have experienced tremendous growth in recent decades. The resolution and quality of electronic imaging has been steadily improving, and digital cameras are becoming ubiquitous. In this respect, Image texture gives us information about the spatial arrangement of color or intensities in an image. Some texture examples are shown in Fig. 1.

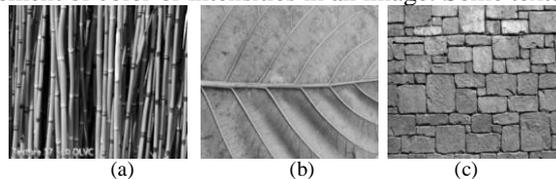

(a)　　　　　(b)　　　　　(c)
**Figure 1.** Some texture examples　(a) Bamboo (c) Leaf (d) Wall

Texture classification task is to associate a class label to its respective category. One of the most efficient texture analysis approaches is local binary patterns (LBP), which provide discriminative features. The LBP is a non-parametric operator which describes the local spatial structure and local contrast of an image. It was described originally for gray-level images, so one of the challenges is how to combine LBP with color sensor information. Sensitivity to impulse noise and low local intensity differences are another big limitations of LBP.

In this paper an approach is proposed for color-texture classification using our proposed version of LBP. First of all, we propose a noise resistant version of LBP which is called Hybrid color local binary patterns (HCLBP). In HCLBP, the separated texture information of color sensors are combined using logical AND operator. Then, a fusion algorithm is used to extract color sensor information and texture features jointly. Also, an algorithm is proposed to select significant LBP to be considered in HCLBP process, which is called Significant Points Selection (SPS) algorithm. SPS decreases computational complexity along with increasing classification accuracy.

In result part, the quality of the proposed approach is evaluated using three data sets: Vistex, Outex, and KTH-TIPS-2a. Our approach is compared with some well known texture classification methods. Our experiments demonstrated that the proposed approach achieves highest classification accuracy in three datasets and different image capturing conditions. Also, results show that using SPS algorithm as preprocess phase, increases classification accuracy. Noise resistant power of proposed HCLBP is compared with previous LBP versions.
Low computational complexity, rotation invariant, multi-resolution, and low noise sensitivity are main advantages of our proposed approach. Also, HCLBP operation is a general method which can be used in many other applications to describe images. The proposed significant points selection algorithm can be used in many other image processing applications to choose key points. The proposed approach can be adapted with output images obtained using every kinds of digital cameras such as single-sensor or three-sensor cameras.

### 1.1. Applications Domain:

Color-texture analysis can be used in many different applications such as:
- *Image retrieval:* Color-Texture features can be used to introduce query and database images.
- *Visual inspection systems (VIS):* Texture analysis can be used to identify fault parts (Smith *et al.*, 2000).
- *Face Recognition:* Texture analysis can be used as pre-processing phase.
- *Visual intelligent sensing systems (VISS):* Extracted information of color sensors in combination with texture features can be used in intelligent sensing systems to achieve image properties (Milella, 2012).
- *Object Tracking:* Texture features can be used to distinguish specific object from background.
- *Medical Image Analysis:* To identify not healthy parts such as tumour in brain images.

### 1.2. Challenges:

Texture classification may be done alone or in combination with other sensed features (e.g. color, shape) to perform the task of recognition. Color-texture classification is not an easy task because of the following challenges:
- Variety of possible Textures
- Conditions under which they are imaged
- Noise effects (Tan and Triggs, 2010)
- Rotation of input texture images (Ojala *et al.*, 2002)

### 1.3. Paper Organization:

The paper is organized as follows: In section2, some related works are discussed which use texture/color features separately/jointly for texture classification. In section 3 theories of some previous versions of LBP are survived. In section 4, our noise resistant color-texture analysis approach is proposed. Also, the section 4 includes significant points selection algorithm. Section 5 presents the experimental results. Finally, the conclusion is presented in section 6.

## 2. RELATED WORKS

Most of texture descriptors apply on grey level images thereby ignoring color sensors information. Therefore, the fusion of color sensor information and texture features may provide discriminative features. The main aim of this paper is to propose a color-texture jointly analysis operation. There exist two strategies namely, early fusion and late fusion, to combine color and texture information. Early fusion algorithms are applied texture descriptors on color planes separately. In this way, finally, a joint color-texture representation is obtained that combines the two cues at the pixel-level. Late fusion combines the two cues at the image level. In this respect, separate histograms are constructed for color and texture. Finally, the two visual cues are combined by concatenating the separate histograms into a single representation (Kan *et al.*, 2014).
In a recent work, discriminative color-texture representation is proposed by Pietikinen *et al.*(2002), which considers color histograms and LBP in an early fusion form. In (Pietikinen *et al.*, 2002), distributions of image colors in RGB and Ohta color spaces were used as features.
Combination of LBP and local color contrast is proposed in (Cusano *et al.*, 2014), as color-texture analysis operation. Their proposed descriptor is invariant with respect to translations of the image plane. In (Kan *et al.*, 2014) a high dimensional representation is proposed as combination of five texture description: binary Gabor patterns, local phase quantization, Weber's law, completed LBP, and binary statistical features. High computational complexity can be mentioned as a big disadvantage of their proposed approach. In (Fekriershad, 2011) a histogram is built based on occurrence probability of some primitive pattern units in color planes separately. Finally, statistical features are extracted from output histogram. A robust descriptor is proposed in (Benco et al., 2014), which is called color-level Co-occurrence matrixes (CLCM). First of all, gray-level co-occurrence matrixes (GLCM) and Gabor filters are used to extract texture features. Then the method is applied in separate planes in the color image. In (Qi *et al.*, 2013), a novel approach is proposed to encode cross-channel texture correlation for color/texture classification task. First, authors quantitatively studied the correlation between different planes using LBP and Shannon's information theory to measure the correlation. Finally, a joint color-texture histogram representation is proposed which is termed cross-channel local binary patterns (CCLBP). Nammalvar *et al.* (2010), proposed an approach which extract color and texture features jointly based on evolutionary genetic algorithm. Nammalvar *et al.*, (2010) uses local binary patterns for texture analysis and k-means clustering to evaluate genetic chrominance plane.

## 3. THEORY OF LCOAL BINARY PATTERNS

### 3.1. Basic Local Binary Patterns (LBP)

LBP is an analysis operator which describes the local contrast and local spatial structure of an image. In order to evaluate the LBP, at a given pixel position $(x_c, y_c)$, LBP is described as an ordered set of binary comparisons of pixel intensities between the center pixel and its neighbors. Neighborhoods could be assumed circular because of achieving the rotation invariant. LBP are defined as follows:

$$LBP_{P,R} = \sum_{k=0}^{P-1} \Omega(f_k - f_c) 2^k \quad (1)$$

Where: $\quad \Omega(x) = \begin{cases} 1 & \text{if } x \geq 0 \\ 0 & \text{else} \end{cases} \quad (2)$

$f_c$ corresponds to the gray value of the center pixel, and $f_k$ to the gray values of the neighbors. P shows the number of neighbors. Fig. 2 shows the process of calculation of LBP code. The $LBP_{P,R}$ operator produces $(2^P)$ different output values, corresponding to the $2^P$ different binary patterns that can be formed.

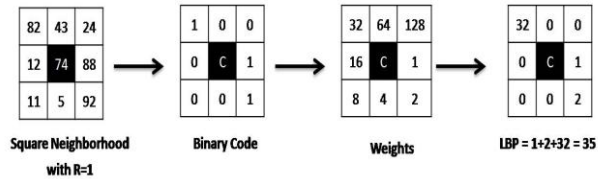

**Figure 2.** LBP Computing Process

When the image is rotated, the gray values $f_p$ will correspondingly move along the perimeter of the circle around $g_c$. To remove rotation effect, a unique identifier to each LBP should be assigned as follows:

$$LBP_{P,R}^{ri} = \min\{ROR(LBP_{P,R}, i) \mid i = 0, 1, \ldots, P-1\} \quad (3)$$

Where ROR(x,i) performs a circular bit-wise right shift on the P-bit number x, *i* times.

### 3.2. Modified LBP (MLBP)

Basic LBP has some disadvantages such as low discrimination and high computational complexity. To solve these problems, Ojala et al., (2002) defined a uniformity measure *"U"*, which corresponds to the number of spatial transitions (bitwise 0/1 changes) in the output pattern. It is shown in Eq. (4). For example, pattern 01001100 have U value of 4, while 11000001 have U value of 2.

$$U(LBP_{P,R}) = |\Omega(f_0 - f_c) - \Omega(f_{p-1}) - f_c| + \sum_{k=1}^{p-1}|\Omega(f_k - f_c) - \Omega(f_{k-1}) - f_c| \quad (4)$$

Patterns with uniformity amount less than $U_T$ are categorized as uniform patterns. The patterns with uniformity more than $U_T$ classified as non-uniforms. Finally, $LBP_{P,R}^{riu_T}$ is computed using Eq. (5).

$$LBP_{P,R}^{riu_T} = \begin{cases} \sum_{k=0}^{P-1} \Omega(f_k - f_c) & \text{if } U(LBP_{P,R}) \leq U_T \\ P+1 & \text{elsewhere} \end{cases} \quad (5)$$

Applying $LBP_{P,R}^{riu_T}$ will assign a label from *0* to *P* to uniform patterns and label *P+1* to non-uniform patterns. In using $LBP_{P,R}^{riu_T}$ just one label P+1 is assigned to all of the non-uniform patterns. To achieve discriminative features, $U_T$ should be optimized that uniform labels cover majority patterns in the image. Experimental results in (Tajeripour and Fekriershad, 2014; Fekriershad and Tajeripour, 2012; Tajeripour *et al.*, 2008) show that if $U_T$ is selected equal to P/4, only a negligible portion of the patterns in the texture takes label P+1. $LBP_{P,R}^{riu_T}$ quantifies the occurrence statistics of individual rotation invariant patterns corresponding to certain micro-features in the image, hence the patterns can be considered as feature detectors. For example, Fig. 3, illustrates the 8 unique rotation invariant $LBP_{8,1}$ that can be occurred. As it is shown in Fig. 3, pattern #0 detects bright spots, #8 dark spots, flat areas, #4 edges, etc.

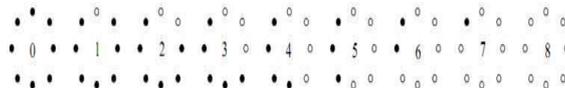

**Figure 3.** The 8 binary patterns that can occur in $LBP_{8,R}$. Black and white circles correspond to bit values of 0 and 1 in the output

As it was described, a label is assigned to each neighborhood. Regarding the Eq. (5), for every image, a feature vector *D* can be extracted as follows:

$$D = <d_0, d_1, \ldots, d_{P+1}> \quad (6)$$

Where:

$$d_k = N_k/N \times M \qquad 0 \leq k \leq P+1 \qquad (7)$$

$N_k$ shows total number of neighbors that labeled as *K*. Also, *N* and *M* are the size of input image. Where, $d_k$ is the occurrence probability of label *k* in whole. To obtain $d_k$, first $LBP_{P,R}^{riu_T}$ should be applied on the whole image and the labels are assigned to neighbors. Then the occurrence probability of each label in the image is regarded as one of the dimensions of the feature vector.

### 3.3. Color-Texture Features Fusion

The basic LBP scheme and most of it's improved versions were defined for grayscale images. We use a simple extension to color images, which is proposed originally in (Pietikainen *et al.*, 2002). Color images can be represented in different color spaces such as RGB, HSV, etc. In this part, the definition is evaluated in RGB. RGB uses an additive color mixing model of three color sensors: red, green and blue colors. To represent color images, separate red, green and blue components must be specified for each pixel and so the pixel value is actually a vector of three numbers, which are known as $f_r$, $f_g$, $f_b$. Often the three different components are stored as three separate `grayscale' images known as *color planes* which have to be recombined when processing. In order to combine color-texture information, each colorful input texture can be separated in three different color spaces. Now, the proposed texture analysis operator (Eq.6) can be computed for each color plans separately(Pietikainen *et al.*, 2002). Finally, the extracted vectors can be concatenated in a simple representation as follows:

$$D = <D_R, D_G, D_B> \qquad (8)$$

Where, $D_R$ shows the extracted feature vector for color plane red using Eq. 6. Also, $D_G$ and $D_B$ can be defined in a similar ways. Finally, vector D has 3P+6 dimensions.

## 4. PROPSOED APPROACH

### 4.1. Proposed Noise Resistant Version of LBP

Image noise is random variation of brightness or color information in images, and is usually an aspect of electronic noise. Noise produces undesirable effects such as artifacts, unrealistic edges, unseen lines, etc. In this respect, different noise models can be defined such as Gaussian, white, Browning, impulse valued, periodic, etc. Impulse noise is seen in data transmission. Some image pixel values are replaced by corrupted pixel values. Impulse Noise sensitivity is one of the big limitations of the LBP, which decreases the analysis accuracy. Even a small noise may change the LBP code significantly. In the evaluation of $LBP_{P,R}^{riu_T}$, threshold is considered at exactly the value of the central pixel, so LBP tends to be sensitive to noise, especially in near uniform image regions. It causes change label of the desired centered pixel which effects on final accuracy. For example, suppose two different local neighbors as it is shown in Fig. 4.a and Fig. 4.b. An impulse noise is applied on Fig.4.a, and output is shown in Fig. 4.c. The extracted labels for each pattern are as follows:
$LBP_{8,1}(a)=8 \quad LBP_{8,1}(b)=0 \quad LBP_{8,1}(c)=0$
The Fig. 4.a shows a local flat area, and Fig.4.b shows a local bright spot. As a human visual inspection result, the Fig. 4.c shows flat area and it is more similar to the Fig. 4.a, but it's LBP label is same with Fig. 4.b.

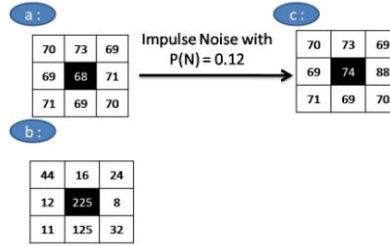

**Figure 4.** Noise effects on MLBP

The texture analysis techniques discussed above had been defined for grayscale images. In a gray-level image like F, impulse noise can be defined as transmission function like N that probability changes some of the image pixel values as follows. Where F' shows the output noisy image.

$$F(x,y) \pm N(x,y) = F'(x,y) \qquad (9)$$

The noise occurrence probability on a specific pixel at given position (x,y), can be considered as P(N). In a colorful image with "i" color planes, at a given pixel position (x, y, i), impulse noise can be defined as follows:

$$F(x,y,i) \pm N(x,y,i) = F'(x,y,i) \qquad (10)$$

We claim that, in a colorful image, when an impulse noise falls on a specific pixel, it may disturb the point intensity in one or more color planes. So, the noise occurrence probability on a specific pixel at given position (x, y, i), can be considered as P(N)/3. In this respect, the probability of occurring noise on all channels of a specific pixel concurrent is very low. In the other words, pixel values of a given pixel at position (x, y) may not change at least in one of the color planes with high probability.

To prove this claim, we evaluated an innovative experiment here to explore the impulse noise effect on color planes. In this respect, 30 texture images where collected using Outex TC_00013 (Ojala *et al.*, 2002), Vistex (Picard *et al.*, 2004), and KTH-TIPS-2a datasets randomly, and an impulse noise was applied on them for 10 times with different ratios. Then, the intensity value of output pixel at given position (x, y) were compared with their intensity value before applying noise. Finally, the average noise effect on *K* color planes is described as follows:

$$Average\ Noise\ effect\ on\ K\ color\ planes = \frac{Number\ of\ Noisy\ pixels\ with\ change\ in\ K\ number\ of\ color\ planes}{Total\ Number\ of\ Noisy\ Pixles} \quad (11)$$

**Table 1:** Average(%) of noise effect on color planes

| Noise Ratio / Noise effect on K color planes | 5 % | 10 % | 20 % | 30 % | 40 % |
|---|---|---|---|---|---|
| Effect on 1 channel | 95.00% | 90.09% | 79.72% | 68.51% | 57.15% |
| Effect on 2 channels | 4.91% | 9.55% | 18.80% | 27.79% | 35.49% |
| Effect on3 channels | 0.07% | 0.34% | 1.46% | 3.69% | 7.34% |

As it is shown in Table1, for example in 0.34% of noisy pixels, the intensity value was changed in all 3 color planes concurrent after applying impulse noise with ratio 10%, which is very low probability. In the other words, impulse noise may disturb the intensity value of a given pixel at position (x, y) in all color planes with very low probability (with common noise ratio in image processing applications or online problems). To assume this mention in texture analysis, function Ω(x) can be new defined as follows:

$$\Omega_X = \begin{cases} 1 & if\ (\Omega_{X_R} \times \Omega_{X_G} \times \Omega_{X_B}) = 1 \\ 0 & if\ (\Omega_{X_R} \times \Omega_{X_G} \times \Omega_{X_B}) = 0 \end{cases} \quad (12)$$

Where,

$$\Omega_{X_i} = \begin{cases} 1 & if\ (f'_{i_k} - f'_{i_c}) > 0 \\ 0 & else \end{cases} \quad (13)$$

Where, $f'_{i_c}$ corresponds to the grey value of the center pixel in color plane *i* (e.g. R, G, and B) in noisy image. $f'_{i_k}$ shows the grey values of the neighborhoods in the same color plane *i* in noisy texture image. An unlike noise on centered pixel or its neighbors in one or two channels, can't change the final pattern and extracted label. In other words, a neighbor pattern can be desired as 1, just when it's intensity value is more than threshold (center value) in all of the color planes. In this respect, a new LBP representation can be defined for color textures with notation HCLBP$_{P,R}$ as abbreviation of Hybrid Color Local binary patterns. For a colorful image in RGB, along with three previous extracted vectors, new feature vector can be extracted using HCLBP$_{P,R}$. Finally, extracted vectors can be concatenated in a single representation as follows:

$$D = <D_R, D_G, D_B, D_H> \quad (14)$$

Where, $D_H$ shows the extracted feature vector based on HCLBP$_{P,R}$. The proposed descriptor is invariant with respect to several transformations in the color spaces that means it can be used in each color spaces.

**4.2. Proposed Multi-Resolution Analysis**

To evaluate HCLBP, a circularly symmetric neighbor set of P pixels placed on a circle of radius R is used. By altering P and R, different operators can be realized for any quantization of the angular space and any spatial resolution. Multi-resolution analysis can be accomplished by combining the information provided by multiple operators varying (P, R). So, the results of choosing the different size of radius (R) and number of neighbors (P), can be concatenating extracted vectors as follows:

$$D_{multi-resolution} = \{Concatenating <HCLBP_{P,R}> \mid R=1,2,3,\ldots;\ P=8,16,32,\ldots..\} \quad (15)$$

**4.3. Proposed Significant Points Selection Algorithm**

In the evaluation of LBP$_{P,R}^{riu_T}$, threshold is considered at exactly the value of the central pixel, so LBP labels tend to be very sensitive to little local neighborhood differences. In the other means, labels are not evaluated meaningful for some local neighbors exactly. For example, as it is shown in Fig.5, LBP(5.a) and LBP(5.b) are same. But, as a human visual see, these neighbors are not same. Fig. 5.a shows a meaningful dark spot and Fig. 5.b is very similar to flat area. Fig.(5.b) is very similar to Fig. (5.c) just with some low difference in terms of intensity, which cannot be detected with human eyes. But, LBP (5.c) is 4, which means a corner area. It is reason of not meaningful difference between center pixel and some surrounded neighborhoods. For example, difference between center pixel and $f_1$ is more meaningful than $f_2$.

In this respect, considering not-meaningful LBPs may decrease the accuracy and increase computational complexity. In this part, an algorithm is proposed to select significant LBP points, which is called Significant Points Selection (SPS).

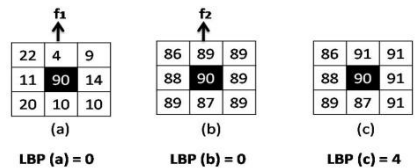

**Figure 5.** significant and not-significant LBP

Local neighborhood differences are not same in different images. For examples, in images with flat background local intensity differences between neighborhoods is very lower than images with different objects in terms of size, shape, color, edge, etc. So, first of all, two measures are proposed to measure local differences in neighborhoods (called Local Significantly Value *(LSV)*), and global average of neighborhood differences (called Global Significantly Value *(GSV)*):

$$LSV = \frac{1}{P} \sum_{k=0}^{P-1} f_k - f_c \quad (16)$$

$$GSV = \frac{1}{N \times M} \sum_{i=1}^{N} \sum_{j=1}^{M} \left( \frac{1}{P} \sum_{k=0}^{P-1} f_k - f_c \right) \qquad (17)$$

Where, N & M show size of image and P is the number of neighbors. Finally each central pixel with LSV more than GSV should be considered as a significant point. Now, SPS can be combined with HCLBP as a preprocess step. It is enough to just consider significant LBPs in computation process of the feature vector D in Eq. 14.

## 5. EXPERIMENTAL RESULTS

To evaluate the performance of our framework, we carried out experiments on three comprehensive color-texture datasets: Outex TC_00013(Oajla *et al.*, 2002), Vistex(Picard *et al.*, 2004), and KTH-TIPS-2a. The database details are shown in Table 2. In our experiments original images of Outex TC_00013 and Vistex were cropped to non-overlap windows to obtain enough samples per class. To evaluate performance on the KTH-TIPS-2a, we use the standard protocol (Sharma *et al.*, 2012) and report the average performance over the 4 runs, where every time all images of one sample are taken for test while the images of the remaining 3 samples are used for training. Some examples of datasets are shown in Fig. 6

**Table 2.** Experimental dataset details

| Dataset | Original Size | Sample Size | Class | Samples Per Class | Challenges |
|---|---|---|---|---|---|
| Outex TC_00013 | 538×746 | 128×128 | 68 | 20 | 3Illuminations, 9 angles |
| Vistex | 512×512 | 128×128 | 54 | 16 | Different viewpoints illumination orientations |
| KTH-TIPS-2a | 200×200 | 200×200 | 11 | 4 | 9 scales, 3 poses , 4 Illuminations |

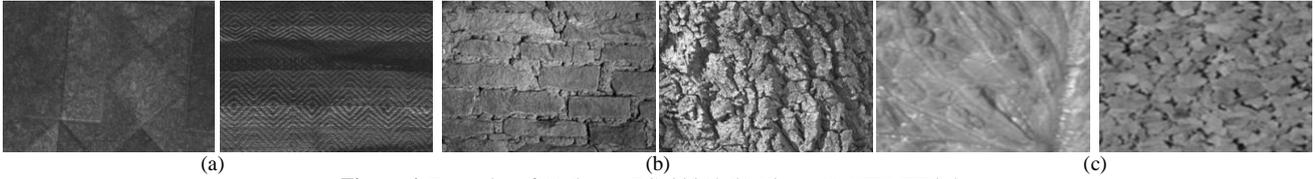

(a)             (b)             (c)
**Figure 6.** Examples of (a) Outex-TC_00013 (b) Vistex (c) KTH-TIPS-2a

### 5.1. Classification Accuracy

Here, we conduct two experiments to validate the effectiveness of the proposed HCLBP and SPS algorithm. In the first experiment, HCLBP is evaluated on three datasets without using SPS algorithm. In the second one, HCLBP is evaluated after applying SPS algorithm on dataset images. To choose most suitable classifier, four different classifiers: Naïve Bayes, Bayesian Network, KNN, and J48 Tree were used. Results are shown in Tables 3-5. Where, 1-NN provides maximum accuracy using multi-resolution $HCLBP_{8,1} + HCLBP_{16,2}$ on Outex and maximum accuracy on Vistex dataset using $HCLBP_{8,1}$. Maximum accuracy was obtained by 1-NN and $HCLBP_{8,1}$ on KTH-TIPS-2a. As it is shown, using SPS as preprocess step increases the classification accuracy in near all of experiments.

**Table 3.** Classification accuracy (%) using different Classifiers on Outex TC_00013 based on 10-fold

| Operator | Bayesian Network | Naïve Bayes | J48 Tree | 1NN | 3NN | 5NN |
|---|---|---|---|---|---|---|
| $HCLBP_{8,1}$ | 92.76 | 94.87 | 86.74 | 99.05 | 98.04 | 96.91 |
| $HCLBP_{16,2}$ | 95.96 | 98 | 87.54 | 99.27 | 98.74 | 97.5 |
| $HCLBP_{8,1} + HCLBP_{16,2}$ | 95.66 | 97.53 | 89.34 | 99.65 | 99.16 | 97.37 |
| $HCLBP_{8,1}$ + SPS | 93 | 95.16 | 87.16 | 99.16 | 98 | 97.16 |
| $HCLBP_{16,2}$+ SPS | 96 | 98 | 87.50 | 99.33 | 99 | 97.5 |
| $HCLBP_{8,1} + HCLBP_{16,2}$+ SPS | 95.66 | 97.66 | 89.50 | **99.83** | 99.16 | 98 |

**Table 4.** Classification accuracy (%) using different Classifiers on Vistex based on 10-fold

| Operator | Bayesian Network | Naïve Bayes | J48 Tree | 1NN | 3NN | 5NN |
|---|---|---|---|---|---|---|
| $HCLBP_{8,1}$ | 93.90 | 94.19 | 89.03 | 97.63 | 97.04 | 95.74 |
| $HCLBP_{16,2}$ | 92.43 | 93.20 | 86.32 | 95.32 | 95.83 | 93.55 |
| $HCLBP_{8,1} + HCLBP_{16,2}$ | 94.52 | 94.12 | 86.32 | 97.83 | 95.73 | 95.1 |
| $HCLBP_{8,1}$+ SPS | 94.16 | 94.16 | 89.37 | **98.12** | 97.08 | 95.83 |
| $HCLBP_{16,2}$+ SPS | 93.12 | 93.33 | 87.5 | 96.45 | 96.04 | 93.95 |
| $(HCLBP_{8,1} + HCLBP_{16,2})$+ SPS | 94.58 | 94.79 | 87.5 | 97.91 | 96.87 | 95.20 |

**Table 5.** Classification accuracy (%) using different Classifiers on KTH-TIPS-2a based on 10-fold

| Operator | Bayesian Network | Naïve Bayes | J48 Tree | 1NN | 3NN | 5NN |
|---|---|---|---|---|---|---|
| $HCLBP_{8,1}$ | 81.97 | 83.11 | 78.31 | 83.57 | 83.31 | 80.9 |
| $HCLBP_{16,2}$ | 82.03 | 81.53 | 74.32 | 83.44 | 82.88 | 79.03 |
| $HCLBP_{8,1} + HCLBP_{16,2}$ | 80.81 | 79.85 | 76.41 | 82.81 | 83.11 | 78.95 |
| $HCLBP_{8,1}$+ SPS | 82.27 | 83.16 | 78.72 | **83.57** | 83.48 | 81.23 |
| $HCLBP_{16,2}$+ SPS | 82.12 | 81.48 | 75.51 | 83.48 | 83.04 | 79.52 |
| $(HCLBP_{8,1} + HCLBP_{16,2})$+ SPS | 80.81 | 79.91 | 76.50 | 82.92 | 83.41 | 80.22 |

Effect of Training Size was studied using following experiment. In a good predictive model the classification rate increases as the number of samples used for training increase. This is indicated in Table 6. .As more training cuts are used for classification, each class is more accurately defined. In these experiments good performance was reached using relatively few training samples. In these experiments we split the Vistex and TC_00013 dataset into training and testing (20% - 80% of samples) without resampling. The multi-resolution $HCLBP_{8,1} + HCLBP_{16,2}$ was used as analysis operator.

**Table 6**. Classification accuracy(%)obtained by resizing train set

| Outex TC-00013 dataset | | | | | | | |
|---|---|---|---|---|---|---|---|
| Train Size (%) Classifier | 20 % | 30 % | 40 % | 50 % | 60 % | 70 % | 80 % |
| 1NN | 93.35 | 96.66 | 96.11 | 97 | 97.91 | 98.88 | 100 |
| 3NN | 90.76 | 93.33 | 93.88 | 95 | 96.25 | 96.11 | 99.84 |
| 5NN | 85.28 | 87.85 | 92.5 | 94.33 | 95.41 | 95.55 | 98.88 |
| Vistex dataset | | | | | | | |
| Train Size (%) Classifier | 20 % | 30 % | 40 % | 50 % | 60 % | 70 % | 80 % |
| 1NN | 93.35 | 92.55 | 94.79 | 95.83 | 95.31 | 95.83 | 97.74 |
| 3NN | 87.54 | 90.47 | 94.09 | 95.41 | 94.79 | 95.13 | 97.36 |
| 5NN | 80.25 | 83.61 | 89.58 | 94.58 | 93.75 | 93.75 | 96.47 |
| KTH-TIPS-2a dataset | | | | | | | |
| Train Size (%) Classifier | 20 % | 30 % | 40 % | 50 % | 60 % | 70 % | 80 % |
| 1NN | 72.65 | 79.35 | 80.75 | 83.57 | 84.31 | 86.21 | 89.41 |
| 3NN | 70.63 | 75.47 | 82.09 | 83.48 | 85.69 | 86.12 | 88.62 |
| 5NN | 70.54 | 74.2 | 80.28 | 81.23 | 81.75 | 83.62 | 87.12 |

## 5.2. Comparison with state-of-the-art

In order to compare effectiveness of our proposed approach, some state-of-the-art methods in this literature are survived. In our experiment to achieve a fair work, four different classifiers: Naïve Bayes, Bayesian Network, KNN, and J48 Tree were used to achieve highest accuracy as follows:

Local Ternary Pattern (LTP) (Tan and Triggs 2010) quantizes local difference into three levels by a threshold. For computation simplicity, the ternary patterns are spitted into two LBPs, positive and negative. Then two histograms are built and concatenated into one histogram. In this experiment, LTP (Tan and Triggs 2010) was applied on color planes to achieve color-texture classification. Finally, $LTP_{16,2}$ provided maximum accuracy in all datasets.

In (Zhenhua *et al.*, 2010), an associated completed LBP algorithm is developed for texture classification. A local region is represented by its center pixel and a local difference sign-magnitude transform. The center pixels represent the image gray level and they are converted into a binary code, namely CLBP-Center (CLBP_C). In our experiment, $CLBP_{P,R}$ was evaluated on three datasets. $CLBP_{24,3}/M_{24,3}/C$ provided maximum accuracy on Outex, however, $CLBP_{16,2}/M_{16,2}/C$ provided maximum accuracy on Vistex and KTH-TIPS-2a datasets.

In MBP (Hafiane *et al.*, 2007), the median gray value of the neighborhood is used instead of center pixel threshold. In order to extend to color images, MBP is evaluated on color planes separately in this experiment. $MBP_{8,1}+MBP_{16,2}$ provided maximum accuracy using 3NN.

As it was described in related works, CLCM (Benco *et al.*, 2014) was evaluated. In order to report maximum accuracy, we uses reported results in (Benco *et al.*, 2014) on Outex and Vistex datasets. The standard proposed approach in (Benco *et al.*, 2014) is implemented on KTH-TIPS-2a.

In Improved local binary patterns (ILBP) (Jin *et al.*, 2004) the threshold used is the mean value of the whole neighborhood including the center pixel. In addition, center pixel will also be a part of the binary code making it $P + 1$ bits long. In order to provide an extended version for color images, ILBP is evaluated on color planes separately in this experiment. $ILBP_{16,2}$ provided maximum accuracy using 3NN.

In (Fekriershad, 2011) a combination of primitive patterns units and statistical features are proposed for texture analysis. Fekriershad (2011) extends primitive pattern units to separate color planes to achieve color-texture classification. In this experiment, 3NN provided maximum accuracy on all three datasets.

Sharma et al., (2012) proposed an approach especially for texture analysis of facial images based on Local Higher Order Statistics (LHS). Our results were same with reported results in (Sharma *et al.*, 2012) on KTH-TIPS-2a. Sharma et al., didn't evaluate LHS on Vistex and Outex. So, we used reported standard evaluation of LHS in (Sharma *et al.*, 2012) to evaluate results in this experiment.

In (Xu *et al.*, 2005), first input image is transformed by wavelet filter. Next, statistical features are extracted from wavelet transformed sub-images. After evaluating reported algorithm in (Xu *et al.*, 2005), 7NN provided maximum accuracy on all three dataset in this experiment.

In (Tajeripour and Fekriershad, 2014), a one dimensional local binary patterns (1DLBP) is proposed where the neighborhood is a row (column) wise line segment. In order to apply 1DLBP, the gray value of the first pixel in the segment is compared with gray value of other pixels in the segment. In this version of LBP, the uniformity measure "*U*" corresponds to the number of spatial transitions (bitwise 0/1 changes) in the row (column) segment. We extended 1DLBP to separate color sensors to achieve color-texture classification. Finally, 1DLBP with row size =8 provided maximum accuracy using 3NN.

As it was described in related works, in (Khan *et al.*, 2014), a compact combination algorithm is proposed based on pure color descriptors and multiple texture descriptors which is known as CCTD.

Ren *et al.*, (2013) proposed an efficient noise resistant version of local binary patterns for facial image analysis. In view of this, NRLBP preserves the image local structures in presence of noise. The small pixel difference is vulnerable to noise. Thus, we encode it as an uncertain state first, and then determine its value based on the other bits of the LBP code. We evaluated reported algorithm in (Ren *et al.*, 2013) to achieve color-texture classification.

The comparison results are shown in Table 7. Results show that our proposed approach provides higher accuracy than other approaches on Outex and KTH-TIPS-2a. In Vistex dataset, MCSFS provides maximum accuracy. The classification rate of MCSFS is 0.89% more than our proposed on vistex. But MCSFS rate is 7.33% lower than our $HCLBP_{P,R}$ on Outex dataset, and 8.04% lower on KTH-TIPS-2a. It shows high mutability of MCSFS, which decreases it's usability in real applications.

**Table 7.** Comparison results (%) of different color-texture representations on three datasets.

| Approach | Outex | Vistex | KTH-TIPS-2a |
|---|---|---|---|
| $LTP_{8,1}$ (Tan and Triggs, 2010) | 98.64 | 96.76 | 71.3 |
| $MLBP_{8,1}$ (Oajal *et al.*, 2002) | 95.83 | 96.04 | 69.8 |
| CCTD (Khan *et al.*, 2014) | 98.20 | 95.52 | 82.7 |
| PPU + SF (Fekriershad, 2011) | 94.10 | 91.37 | 71.28 |
| CLCM (Benco *et al.*, 2014) | 93.27 | 92.57 | 72.15 |
| ILBP (Jin *et al.*, 2004) | 91.26 | 91.47 | 72.68 |
| MBP (Hafiane *et al.*, 2007) | 92.37 | 93.59 | 74.62 |
| $CLBP\_S_{P,R}/M_{P,R}/C$ (Zhenhua *et al.*, 2010) | 98.33 | 95.93 | 73.1 |
| MCSFS (Porebski *et al.*, 2013) | 92.5 | **98.8** | 74.61 |
| Wavelet Features (Xu *et al.*, 2005) | 85.2 | 86.43 | 63.83 |
| 1DLBP (Tajeripour and Fekri ershad, 2014) | 97.9 | 94.9 | 79.75 |
| LHS (Sharma *et al.*, 2012) | 97.32 | 93.26 | 73.0 |
| NRLBP (Ren *et al.*, 2013) | 97.23 | 94.67 | 78.32 |
| DMD (Mehta *et al.*, 2016) | 97.14 | 94.53 | 78.5 |
| $HCLBP_{8,1}$ | **99.65** | 97.83 | **83.57** |
| $HCLBP_{8,1}$ + SPS | **99.83** | 98.12 | **83.57** |

### 5.3. Noise Resistance

To compare the noise resistance power of our proposed HCLBP, An impulse noise (salt & pepper) was applied on database images with different ratios. Some of the noisy images are shown in Fig. 7. Then LTP (Tan and Triggs, 2010), MLBP (Oajal *et al.*, 2002), MBP (Hafiane *et al.*, 2007), CLBP (Zhenhua *et al.*, 2010), and NRLBP(Ren *et al.*, 2013) were evaluated. Finally, classification accuracy was evaluated using 10-fold and 1NN on above three datasets to have a fair experiment. Results are shown in Table 8. After applying noise, $HCLBP_{P,R}$+SPS provides higher classification accuracy than previous versions of LBP in near all of the experiments, which shows it's impulse noise resistance.

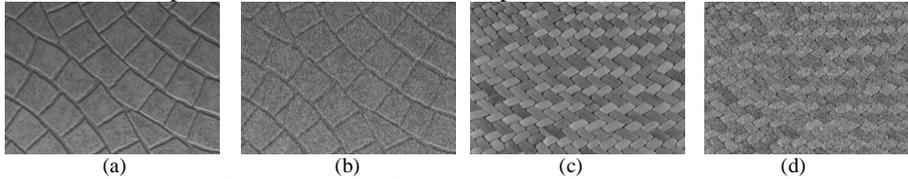

(a)　　　　　(b)　　　　　(c)　　　　　(d)
**Fig 7.** Some examples of corrupted images after applying noise
(a) Original Image (b) Corrupted image with noise 30%　(c) Original Image (d) Corrupted image with noise 30%

**Table 8.** Classification accuracy (%) of different versions of LBP on three texture datasets after applying impulse noise with different noise ratio

| Dataset | Outex | | | | Vistex | | | | KTH-TIPS-2a | | | |
|---|---|---|---|---|---|---|---|---|---|---|---|---|
| Noise Ratio / Approach | 5% | 15% | 20% | 30% | 5% | 15% | 20% | 30% | 5% | 15% | 20% | 30% |
| $MLBP_{8,1}$ (Ojala et al., 2002) | 90.25 | 70.66 | 68.35 | 62.33 | 91.81 | 81.87 | 72.46 | 67.95 | 69.24 | 67.32 | 60.24 | 57.31 |
| $LTP_{8,1}$ (Tan and Triggs, 2010) | 96.24 | 92.36 | 87.54 | 83.47 | 94.31 | 90.64 | 86.25 | 82.31 | 70.93 | 70.45 | 68.32 | 62.35 |
| $CLBP\_S_{8,1}/M_{8,1}/C$ (Zhenhua *et al.*, 2010) | 97.41 | 95.26 | 90.15 | 87.54 | 93.27 | 90.16 | 89.17 | 88.27 | 73.0 | 72.54 | 70.43 | 68.42 |
| MBP (Hafiane *et al.*, 2007) | 91.27 | 90.63 | 86.45 | 76.54 | 92.26 | 90.54 | 88.31 | 82.37 | 70.46 | 68.14 | 63.27 | 59.76 |
| NRLBP (Ren *et al.*, 2013) | 94.53 | 92.42 | 91.50 | 85.36 | 95.31 | 94.9 | **94.73** | 90.51 | 81.26 | 80.42 | 79.62 | 70.53 |
| $HCLBP_{8,1}$ + SPS | **98.94** | **96.5** | **93.34** | **87.5** | **97.84** | **97.5** | 94.62 | **93.95** | **83.42** | **81.53** | **80.14** | **72.42** |

### 5.4. Mathematical/Computational Complexity

In order to compute complexity, one of the efficient methods is computing total number of required mathematical/logical operations such as addition, subtraction, multiplication, division and comparison (Tajeripour and Fekriershad, 2014; Tajeripour *et al.*, 2008). In our proposed approach, the total numbers of required operations, which are applied on each test image, are related to the size of image and size of HCLBP. If image sample size considered as W×W, so the total number of HCLBP operations that are applied on each sample is equal to (W-2R) × (W-2R). Where, R shows the radius size $HCLBP_{P,R}$. Finally, the number of required mathematical operations to apply one $HCLBP_{P,R}$ operation should be computed. The total number of required operations for applying $HCLBP_{P,R}$ with different R and P in a sample image in size of 128×128 are shown in Table 9.

Table 9. Number of required operations evaluating HCLBP

| Operator | Comparison | Multiplication | Division | Addition | Subtraction |
|---|---|---|---|---|---|
| HCLBP$_{8,1}$ | 127008 | 254016 | 0 | 127008 | 127008 |
| MLBP$_{8,1}$ | 127008 | 0 | 12708 | 127008 | 127008 |
| HCLBP$_{8,1}$+SPS | 82560 | 165120 | 0 | 82560 | 82560 |

In (Tajeripour and Fekriershad, 2014; Tajeripour *et al.*, 2008) authors proved that computational complexity of MLBP is less than many other LBP versions and texture analysis methods. As it is shown in Table 9, HCLBP just uses $(2P)\times[(W-2R)\times(W-2R)]$, Multiplication operations more than MLBP, which is not too much. Also, statistical experiment shows that using SPS algorithm decreases the total number of required operations because of eliminating not-significant points (about 35% of HCLBP points). In this respect, computational complexity of the HCLBP+SPS is absolutely lower than many other LBP versions.

## 6. Conclusion

In this paper a noise resistant and multi-resolution version of LBP, that is called HCLBP, was proposed. HCLBP extracts color-texture features jointly. Significant points selection algorithm was proposed as a preprocess step to select most significant LBPs. The performance of our proposed approach was evaluated on three popular datasets. Comparison results demonstrated high classification accuracy and low impulse noise sensitivity of our proposed operation in comparison with state-of-the-art methods.

Results show that using SPS algorithm can decrease the computational complexity along with increasing accuracy. Rotation invariant and low computational complexity are some other main advantages of the proposed approach, which make it useful for online applications. The proposed HCLBP is a general operation to extract color/texture features jointly, which can be used in many other image processing applications. Also, proposed SPS has a general theory, which makes it useful in other cases to select image key points. One of the big advantages of proposed approach is adapting with output images obtained using every kinds of digital cameras such as single-sensor or three-sensor cameras.